%% file: main.tex
\documentclass{article}
\usepackage{iclr2027_conference,times}
\iclrfinalcopy

\usepackage{hyperref}
\usepackage{url}
\usepackage{booktabs}
\usepackage{graphicx}
\usepackage{amsmath}
\usepackage{amssymb}
\usepackage{multirow}
\usepackage[most]{tcolorbox}
\usepackage{float}
\usepackage{tabularx}
\usepackage{array}
\usepackage{marvosym}

\newlength{\sbcap}
\newlength{\halffigheight}
\definecolor{cardbg}{HTML}{F3F0FA}
\definecolor{cardaccent}{HTML}{4B3C8C}
\newtcolorbox{titlecard}{
  enhanced,
  frame hidden,
  arc=10pt,
  colback=cardbg,
  boxsep=0pt,
  left=0.7cm, right=0.7cm, top=0.5cm, bottom=0.45cm,
  before skip=0pt, after skip=0.4cm,
}

\hypersetup{
  pdftitle={EnvLunch: The Environment as a Free Lunch for Learning to Interact},
  pdfauthor={Wei Liu, Min Gao, Yali Du, Yulan He},
}

\newcommand{\ours}{\textit{EnvLunch}}

\begin{document}

\lhead{Preprint}

\begin{titlecard}
\begin{center}
{\LARGE\bfseries\color{cardaccent} \textit{EnvLunch}: The Environment as a Free Lunch\\[2pt] for Learning to Interact\par}
\vspace{0.35cm}
{\normalsize
 \textbf{Wei Liu}\textsuperscript{$\spadesuit$\,\Letter},\quad
 \textbf{Min Gao}\textsuperscript{$\diamondsuit$},\quad
 \textbf{Yali Du}\textsuperscript{$\spadesuit\clubsuit$},\quad
 \textbf{Yulan He}\textsuperscript{$\spadesuit\clubsuit$\,\Letter}\par}
\vspace{0.12cm}
{\small \textsuperscript{$\spadesuit$}King's College London \quad \textsuperscript{$\diamondsuit$}Chongqing University \quad \textsuperscript{$\clubsuit$}The Alan Turing Institute\par}
\end{center}
\vspace{0.12cm}
\noindent{\color{cardaccent!45}\rule{\linewidth}{0.5pt}}
\vspace{0.08cm}

{\small\noindent
\textbf{\color{cardaccent}Abstract.}~For agentic data analysis, language-model agents often require a large number of annotated tasks to learn how to interact effectively with executable environments (i.e., databases). While high-quality annotated tasks require heavy manual effort, an executable environment already provides rich interaction structure and verifiable outcomes. In this sense, the environment itself offers a free lunch, an additional set of verifiable tasks that require no human annotation. We propose \ours{}, a warm-up stage that turns this environmental signal into agent training without human-written or model-generated task descriptions, before downstream agentic reinforcement learning. \ours{} samples valid action trajectories, executes them to obtain final observations, and trains an agent to recover a functionally equivalent trajectory given only its final observation. This label-free warm-up enables the model to outperform the baseline by over $9$\% on average across all downstream training budgets, and with few labels, the baseline needs $8$ to $32$ times as many annotated examples to match the accuracy that the \ours{} model reaches at its smallest training budget. Letting the environment alone define the task yields a larger gain than having the model write its own tasks over the same environments, and this warm-up also improves fine-tuning on benchmarks beyond data analysis. \ours{} therefore saves annotated tasks and provides a better starting point for fine-tuning agentic models when such tasks are limited.\par}
\vspace{8pt}

\noindent{\footnotesize\textbf{Correspondence:}~\texttt{\{wei.4.liu, yulan.he\}@kcl.ac.uk}}\hfill
\raisebox{1.5pt}{\includegraphics[height=0.58cm]{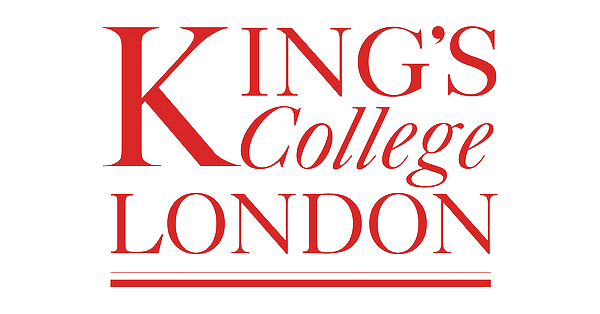}}\hspace{0.25cm}\raisebox{1.5pt}{\includegraphics[height=0.55cm]{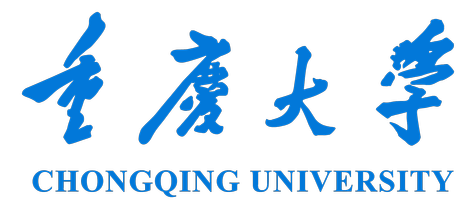}}\hspace{0.25cm}\raisebox{3pt}{\includegraphics[height=0.48cm]{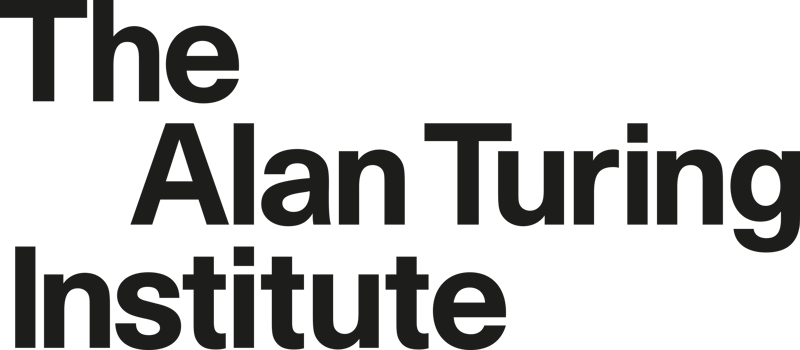}}%
\end{titlecard}

{\renewcommand{\thefootnote}{}\footnotetext{Code and data: \url{https://github.com/thinkwee/EnvLunch}}}

\begin{figure}[h]
\centering
\includegraphics[width=\linewidth]{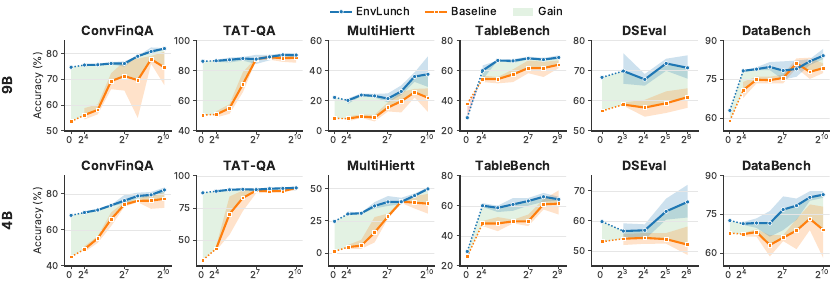}
\caption{Label-efficiency curves for agentic data analysis. The top row shows the $9$B model and the bottom row the $4$B model, with one benchmark per column. The x-axis is the number of annotated examples on a log scale, and $0$ denotes evaluation without downstream fine-tuning. Lines show the mean over three runs, and shaded bands span the corresponding min--max. The green areas between the lines are the gain from the warm-up; red regions indicate the reverse.}
\label{fig:main}
\end{figure}

\section{Introduction}

Language model agents can interact with databases to perform long-horizon and complex data querying, analysis, and report generation \citep{zhang2024dseval,oses2024databench,wu2024tablebench,lei2024spider2}, a setting we refer to as \textbf{agentic data analysis}. Training agents to acquire such capabilities currently relies heavily on either large-scale collections of human-annotated verifiable tasks for reinforcement learning \citep{bai2024digirl,jin2025searchr1,xi2025agentgymrl,ma2025sqlr1,zhang2025agenticrlsurvey}, or long-horizon trajectories distilled from much larger models \citep{sun2025osgenesis,su2025learnbyinteract,tongyi2025deepresearch}. Both approaches require substantial human effort and careful data curation \citep{huang2025envscalingsurvey}.

Yet agentic training has an abundant source of supervision that is often underused, which is the \textbf{executable environment itself} \citep{yu2018spider,li2023bird,lei2024spider2}. In data analysis, databases and tables expose rich state and action spaces, while execution provides an automatic signal for the outcome of an action sequence \citep{tran2009qbo,tran2014qre,wang2017scythe,silver2025era,yao2025secondhalf,xi2024agentgym,liu2025gem,openenv2026}. Existing methods partly reduce annotation costs by synthesising tasks with LLMs or human-defined templates, but still require these environments to be converted into explicit questions or task statements \citep{sun2025osgenesis,zhou2025selfchallenging,stojanovski2025reasoninggym}. We ask whether this step is necessary at all: can an agent learn to interact effectively by learning directly from information inherent in the environment?
Any executable trajectory in an environment could expose the interaction process, providing signals for interaction behaviours such as recovering from failures or exploring the environment over long-horizon interactions. Such trajectories form an additional set of verifiable tasks without human annotation, which we call the free lunch of the environment.

Therefore, we propose \ours{}, an environment-driven warm-up stage that runs before downstream agentic reinforcement learning. \ours{} randomly composes valid actions into an executable trajectory, executes it to obtain a final observation, and trains the agent to recover a trajectory that could have produced that observation. Intuitively, this is like placing a target at a random location in a maze and training the agent to navigate to it: the target need not correspond to a meaningful downstream question, as long as reaching it requires effective interaction with the environment. To prevent the agent from simply copying the target observation, we create multiple perturbed copies of the original environment and execute the recovered trajectory in each, and agreement across these counterfactual environments determines the reward. \ours{} requires no human or model annotation, is fully verifiable, and is lightweight, taking $20$ minutes on CPU to construct $1600$ warm-up tasks compared with more than $4$ hours for model-based task synthesis.

We validate \ours{} on Qwen3.5-4B and Qwen3.5-9B across six agentic data-analysis tasks. Averaged across different training budgets, \ours{} improves accuracy by $9.6$\% at 4B and $9.3$\% at 9B, with the largest gains when labels are scarce, where the baseline requires 8--32$\times$ as many labelled examples to match the accuracy reached by the \ours{} model at its smallest training budget. Environment-driven warm-up also yields larger gains than reinforcement learning on tasks that the model synthesises over the same environments. Across seven additional agentic tasks spanning text-to-SQL, question answering, software use, and embodied interaction, \ours{} also improves downstream performance, despite using warm-up environments disjoint from the evaluation tasks. Finally, in a constructed environment where uncertainty can be computed exactly, \ours{} produces the largest reduction in uncertainty about the environment, providing evidence that learning to interact contributes to the downstream gains.

Our contributions are threefold. First, we show that an executable environment alone can define verifiable agentic tasks without any annotation. Second, we propose \ours{}, a warm-up stage for agents to learn from the environment without human annotation. Third, across two model scales and thirteen benchmarks, we show that \ours{} substantially reduces the annotated examples needed for downstream agentic reinforcement learning, which suggests that executable environments contain useful training signals that can be extracted without first specifying the meaningful tasks. \ours{} provides a preliminary step towards agentic training that learns directly from environments.

\section{Related work}
\label{sec:related}

\paragraph{Learning from executable environments.}
Rewarding a model by executing its output originated in semantic parsing and program synthesis \citep{zhong2017seq2sql,le2022coderl,gehring2025rlef} and now extends to agents operating over web pages, devices, and more \citep{qi2024webrl,bai2024digirl,jin2025searchr1,wei2025swerl,pan2024sweGym,qian2025toolrl,zhang2025agenticrlsurvey}. Executable environments thus act as verifiers that require neither rubrics nor annotation, which has motivated a shift from learning on text towards learning from experience \citep{silver2025era,yao2025secondhalf}, RL infrastructure for existing environments \citep{xi2024agentgym,xi2025agentgymrl,liu2025gem,openenv2026}, and pipelines that synthesise new ones \citep{song2026envscaler,xu2026envfactory,dong2026agentworld,tu2026scaleenv,wang2026agentworldmodel,gandhi2026endlessterminals,hu2024agentgen,guo2025genenv,huang2025envscalingsurvey}, including hand-crafted, model-generated or mined from the web \citep{zeng2026rlve,shi2026evoenv,fang2026agentscaler,cai2025autoforge,zhang2026infiniteweb,verma2025ggbench}. \ours{} builds on this view and carries it one step further: an environment that already supplies the verdict can also supply the task. Any environment these pipelines produce can therefore feed \ours{} directly.
A growing body of work lets the environment shape the tasks as well. Agents explore an environment and turn the resulting trajectory into an instruction \citep{sun2025osgenesis,su2025learnbyinteract,murty2024nnetnav,wang2026trajectory2task,mai2025cues,yang2026coevolve}, a proposer and a solver improve through self-play \citep{zhao2025absolutezero,zhou2025selfchallenging,zhai2025agentevolver,lu2025searchselfplay,huang2025rzero}, task generators are paired with verifiers \citep{stojanovski2025reasoninggym,li2025internbootcamp,chen2025enigmata,sullivan2025randomworld,yang2025sweSmith,jain2025r2egym}, and agents learn from the states their own actions reach along expert demonstrations \citep{zhang2025earlyexperience}. Together these works show that interaction with an environment is a rich source of training signal when little is annotated. \ours{} follows this direction to the point where the environment defines the task and judges it, so no question needs to be written, and even a trajectory with no semantic meaning trains the multi-turn interaction behaviours that downstream tasks require.

\paragraph{Recovering a query from its output.}
Recovering a query from its output is a classic problem in databases \citep{tran2009qbo,tran2014qre,zhang2013reverse,tan2018regalplus,wang2017scythe,takenouchi2021patsql,orvalho2020squares,pradhan2025xpose}, which establishes that an output constrains the program that produced it. This research identifies \emph{instance-equivalent} programs, which agree on one database but compute different functions, and separates them with equivalence provers \citep{he2024verieql}, language model judges \citep{zhao2023llmsqlsolver}, or additional databases designed to distinguish candidate programs \citep{chandra2015datagen,chen2025parseval,klopfenstein2025spotit}. Execution-based text-to-SQL training meets the same issue as false positives of single-instance matching \citep{ma2025sqlr1,pourreza2025reasoningsql,yao2025arctictext2sql,zhu2026revisql}, and synthetic-query training sidesteps it by providing the query explicitly \citep{liu2022tapex,yu2021grappa,li2025omnisql,zhang2025exesql}. \ours{} turns this recovery problem into a training task for agents. Any trajectory that reproduces the target counts as a solution, which makes a task with no statement trainable, and the perturbed copies of Section~\ref{sec:method} rule out degenerate solutions such as copying the output. Hindsight relabelling in goal-conditioned reinforcement learning shows that reached outcomes make effective goals \citep{andrychowicz2017her,ghosh2021gcsl}. \ours{} brings this principle to language-model agents and uses the outcomes of trajectories the agent never sees as the tasks themselves to bring the pressure of environment interaction.

\section{Method}
\label{sec:method}

\begin{figure}
    \centering
    \includegraphics[width=0.99\linewidth]{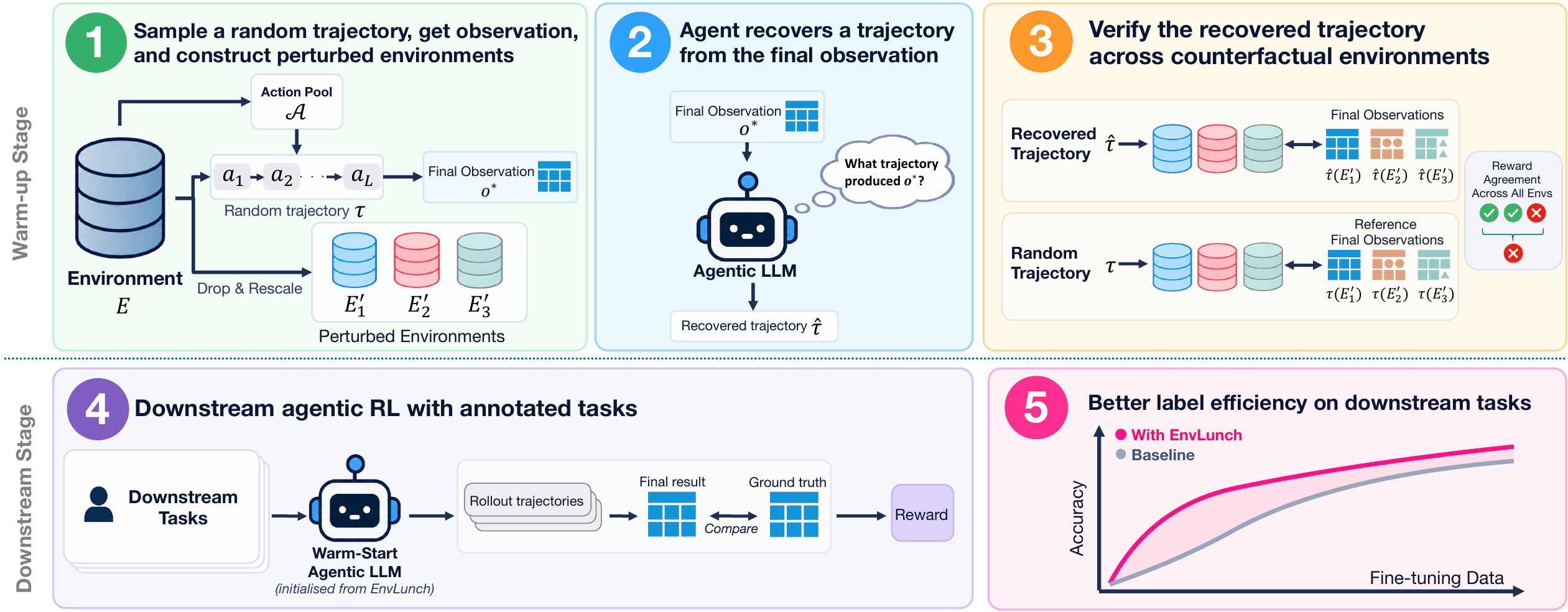}
    \caption{The overview of \ours{}. In stages 1--3, \ours{} performs warm-up by training the agent to recover a random
    synthesised trajectory in the environment.
    Stages 4--5 are ordinary downstream agentic RL, which then needs fewer annotated tasks.}
    \label{fig:framework}
\end{figure}

The free lunch in \ours{} is the additional set of verifiable tasks that the environment provides without human annotation. Any executable trajectory can define a task, and its execution result is the answer, so neither a question nor a manually annotated label is needed. In this sense, the environment itself provides the task where agentic reinforcement learning can be performed.
For agentic data analysis, the database serves as the environment, while database queries and computation statements constitute the executable action space. By constructing an action pool and performing online trajectory synthesis and filtering, and training the agent through counterfactual multi-environment verification (Figure~\ref{fig:framework}), \ours{} provides a fully automated warm-up for agentic capabilities.

\paragraph{Action Pool Construction.}
We define the action space associated with each environment as $\mathcal{A}$, e.g., \texttt{select}, \texttt{where}, and \texttt{group by} in data analysis. These operators form a generic interface
that applies to any relational database without per-environment design. They are further organised into different levels according to the number of operators composed. On the SQL side, the pool has fifteen operators: ten aggregate at increasing levels of composition, from a grouped aggregate up to a three-table foreign-key chain or a common table expression compared against its own average, and five do not aggregate, such as a projection under a filter or a top-$k$ ranking with a tie-break. On the \texttt{pandas} side, a trajectory composes one source action (a table, or two tables merged on a foreign key), a chain of $0$, $1$--$2$, $3$--$4$ or $5$--$6$ transforms depending on the level (filters, top-$k$, grouped aggregates, column ratios or ranks), with a join added at the two highest levels with probability one half, and one terminal action such as a scalar aggregate, a quantile or a row count. The actions can be composed sequentially and hierarchically into a trajectory. We write $E$ for the environment, consisting of a database instance together with its tables, and $\mathcal{A}(o)$ for the actions that are valid given an observation $o$. Action parameters are instantiated dynamically from each environment; for example, the selected fields or number of retrieved rows are sampled according to the underlying database instance.
Let $E \sim \mathcal{D}_E$ denote an environment sampled from the environment distribution, and let $o_0$ be the initial observation it exposes.
Given $E$, \ours{} composes a trajectory by repeatedly drawing a valid action and executing it:
\begin{equation}
\label{eq:synth}
\tau = (a_1,\ldots,a_L),
\qquad
a_\ell \in \mathcal{A}(o_{\ell-1}),
\qquad
o_\ell = a_\ell(o_{\ell-1}),
\end{equation}
where each $a_\ell$ is drawn uniformly at random from the actions that are valid under the current observation, and $a_\ell(o_{\ell-1})$ denotes the observation obtained by executing $a_\ell$ on $o_{\ell-1}$. Since every action is drawn from those valid on the observation it will be applied to, $\tau$ is executable by construction. The length $L$ is drawn from the difficulty levels of the pool and cut short whenever no valid action remains. For example, an \emph{aggregation} or \emph{join} action applied to $o_{\ell-1}$ yields a new observation $o_\ell$. Executing the full trajectory produces the target observation
$
o^\star = o_L = \tau(E).
$
\ours{} reveals only $E$ and $o^\star$, while hiding the sampled trajectory $\tau$. The warm-up task is therefore
\begin{equation}
\mathcal{T}(E,\tau)=(E,o^\star).
\end{equation}
The agent is asked to interact with $E$ to discover the trajectory that could reproduce the target,
\begin{equation}
\label{eq:equiv}
\hat{\tau} \in [\tau]_E := \left\{\tau' : \tau'(E) = o^\star\right\},
\end{equation}
so that the supervision is the execution result rather than a match against the hidden $\tau$. Because the map from trajectories to observations is many-to-one, this class contains functionally equivalent trajectories that the synthesiser never sampled, and the counterfactual verification below is what removes its degenerate members.
Note that recovering a short synthetic trajectory may require substantially longer interaction with the environment.
A constructed trajectory of length $L = 5$, for example, may require many more turns of trial and reasoning than its length, providing substantially more opportunities for learning (e.g., the $14$ actions in Figure~\ref{fig:case}(a)).

\paragraph{Online Trajectory Synthesis and Filtering.}
Each time a new action is appended to the constructed trajectory, we execute the partial trajectory in the environment to obtain $o_\ell$, and rebuild the candidate pool $\mathcal{A}(o_\ell)$ for the next step, keeping an action only if its operands exist in the current observation and it can still change it%
. Its parameters are read off that observation, so a threshold is an actual quantile of the column and a category filter names a value the column holds. For example, once the current execution has left only a handful of rows, a top-$k$ ranking action no longer changes the observation and is therefore removed from the pool.
Such synthetic trajectories need not resemble useful user tasks%
, for example, \textit{``Pair every postseason batting record of a player with his every postseason fielding record, then group those pairs by the fielding error count and report how many pairs fall in each group.''} Such tasks may have little practical meaning,
but they still require the agent to interact with the environment and reason over intermediate observations to recover an executable trajectory.
In addition to dynamically adjusting the candidate action pool, we also check whether the trajectory produces different results when executed under counterfactual perturbations of the environment. We retain a trajectory only if it executes and yields a non-degenerate result in every perturbed copy, and its result changes in at least one copy, so that an answer which merely copies the target observation cannot obtain the full reward.
This rules out the reward hacking illustrated in Figure~\ref{fig:case}(a).

\paragraph{Counterfactual Multi-Environment Verification.}
After obtaining a synthetic trajectory and its final execution result, we use standard agentic RL to train the agent to recover a functionally equivalent trajectory. The agent is provided with the final execution result, the corresponding database instance, SQL and Python tools, and a minimal ReAct harness~\citep{yao2023react}. Let $\hat{\tau}$ denote the trajectory produced by the agent, and $\hat{\tau}(E)$ its execution result in environment $E$, following the notation of Equation~\ref{eq:synth}. We do not reward exact recovery of the original trajectory,
because different trajectories can represent the same logic. Instead, we
compare the execution results of the two trajectories. However, this setup is vulnerable to a simple form of reward hacking, where the agent could simply print the target observation result rather than reconstructing the trajectory. To address this, \ours{} adopts counterfactual multi-environment verification. We perturb the original database by deleting rows and rescaling numeric columns%
, to construct $K$ parallel ``what-if'' databases $E'_1, \dots, E'_K$, which the agent never sees. Each copy deletes $30\%$ of the rows of every table with at least twenty rows, cascading along foreign keys so that the copy stays referentially consistent; on the dataframe side, every measure column is also rescaled by a random factor, while identifier and key columns are never changed. We use $K = 2$ copies for SQL and $K = 3$ for \texttt{pandas}. This replaces the equivalence class of Equation~\ref{eq:equiv} by its intersection over the copies, $\bigcap_{k=1}^{K}\{\tau' : \tau'(E'_k) = \tau(E'_k)\}$, which a trajectory that merely prints $o^\star$ no longer belongs to.
Without the copies, the agent hacks the reward and stops querying the database, all of its recovered queries become hardcoded constants, and accuracy on $208$ held-out warm-up tasks falls to $16.8$, below $28.1$ for the untrained model and $46.1$ with the copies. We execute both trajectories in each copy and reward agreement across all perturbations:
\begin{equation}
\label{eq:reward}
R(\hat{\tau};\tau,E'_{1:K}) \;=\; \Big(1 + w\,(1 - t/t_{\max})\Big)\;\cdot\; \min_{1 \le k \le K}\, m\!\left(\hat{\tau}(E'_k),\, \tau(E'_k)\right),
\end{equation}
where $m \in [0,1]$ measures agreement between two execution results and equals $1$ only when they agree exactly; it is the overlap of row multisets for relational results and an exact comparison after canonicalisation for dataframe results. The leading factor rewards brevity: $t$ is the number of turns the agent spends interacting with the environment before submitting $\hat{\tau}$, $t_{\max}$ is the turn budget of the harness ($16$ for SQL and $48$ for \texttt{pandas}), and $w = 0.5$ weights this bonus. Note that $t$ counts the agent's own interaction and is typically far larger than the length $L$ of the hidden trajectory $\tau$. The min operation requires the recovered trajectory to agree with all $K$ counterfactual environments.

Formally, given the environment $E$ and the target observation $o^\star$, the agent interacts with the environment for multiple turns in a rollout and submits a recovered trajectory $\hat{\tau}$ in its final action
\begin{equation}
\hat{\tau} \sim \pi_\theta(\cdot \mid E,o^\star).
\end{equation}
The \ours{} objective is to maximise the expected reward of the recovered trajectory:
\begin{equation}
\max_{\theta}\;\mathbb{E}_{\substack{E\sim\mathcal{D}_E,\;
\tau\sim\operatorname{Synth}(E),\\
\hat{\tau}\sim\pi_\theta(\cdot\mid E,o^\star)}}
\left[
R(\hat{\tau};\tau,E'_{1:K})\right],
\end{equation}
where $\mathcal{D}_E$ denotes the distribution of environments that could be used for \ours{} warm-up, $\operatorname{Synth}(E)$ is the randomised synthesiser of Equation~\ref{eq:synth}, $o^\star=\tau(E)$ is the target observation it produces, and $E'_1,\ldots,E'_K$ are counterfactual perturbations of the original environment. This objective encourages the agent to recover not necessarily the original action sequence, but a trajectory that implements the same computation under changes to the environment. We optimise the model using standard GRPO~\citep{shao2024deepseekmath}.

\section{Experimental setup}
\label{sec:setup}

We evaluate \ours{} on Qwen3.5-4B and Qwen3.5-9B~\citep{qwen3.5}. For each backbone, \ours{} uses the same pool of environments and warms up only once for every downstream benchmark.
For each benchmark, we fine-tune both the released backbone checkpoint, which we call the \textbf{baseline}, and the corresponding \ours{}-warmed checkpoint, which we call the \textbf{\ours{} model}, using GRPO under the benchmark's native multi-turn interaction protocol. We leave its action space, turn limit, trajectory token budget, and official scorer unchanged. Each downstream fine-tuning setting is repeated with three independent random seeds. We plot downstream accuracy against the number of annotated training examples, evaluating the checkpoints at which the number of training examples seen is a power of two, to evaluate whether \ours{} can improve downstream label efficiency. \ours{} uses
two environment corpora. The first is a relational corpus built on $112$ databases from the training splits of Spider \citep{yu2018spider}, BIRD \citep{li2023bird} and Archer \citep{zheng2024archer} ($63$ from Spider, including the $7$ Archer databases, and $49$ from BIRD), on which the agent interacts through SQL. The second is a dataframe corpus containing $398$ tables from DataMind \citep{qiao2025datamind}, BIRD and DABench \citep{hu2024dabench} ($351$ reinforcement-learning training tables from DataMind, $32$ BIRD training databases read as dataframes, and $15$ DABench tables), on which the agent interacts through Python. All warm-up tasks are built on these two corpora.
The environments are not constructed for any particular downstream benchmark, and no evaluation environment from any benchmark is included in the warm-up data: the builder excludes by name the development databases of Spider and BIRD and every table in the DABench evaluation manifests. Since \ours{} needs no annotation, it can also run on the environments of a target benchmark, but we keep the two disjoint so that the downstream gains measure what transfers across environments rather than familiarity with the evaluation ones.

LLMs following the warm-up stage using \ours{} are then further fine-tuned and evaluated on six downstream data-analysis benchmarks. Four present tables either alone or interleaved with text (ConvFinQA, \citealp{chen2022convfinqa}; TAT-QA, \citealp{zhu2021tatqa}; MultiHiertt, \citealp{zhao2022multihiertt}; TableBench, \citealp{wu2024tablebench}). The remaining two, DSEval \citep{zhang2024dseval} and DataBench \citep{oses2024databench}, require interacting with dataframes through code. Each keeps its own official test set and metric, which is proportional match except for the official DataBench accuracy. The held-out sizes and turn limits are $1486$ and $16$ for ConvFinQA, $494$ and $12$ for TAT-QA, $300$ and $32$ for MultiHiertt, $118$ and $16$ for TableBench, $87$ and $20$ for DSEval, and $351$ and $16$ for DataBench. The fine-tuning pool of a benchmark is drawn from its own training split and subsampled to at most $2048$ examples, and training budgets range over $16$--$1024$ examples ($16$--$512$ for TableBench and $8$--$64$ for DSEval). For benchmarks without a usable training split (DSEval, and Spider2-lite and SpreadsheetBench in Section~\ref{sec:cross}), we split the data at the level of the underlying source (Spider2-lite by database) with disjoint identifiers, which leaves pools of $99$, $74$ and $493$ examples.
We follow the official settings as closely as possible and add no extra harness design or result ensembling for chasing SOTA, since the quantity of interest is downstream label efficiency. The warm-up runs $100$ steps of $16$ tasks each, which takes about $4$ hours for the $9$B model and $3$ hours for the $4$B model. The warm-up needs no annotated tasks, runs once for each backbone and serves all downstream benchmarks. We also conduct experiments to rule out the possibility that the observed effect simply results from additional training on synthetic data (Section~\ref{sec:subject}).

\paragraph{Training details.}
We train with slime \citep{slime_github}, which runs Megatron-LM \citep{shoeybi2019megatron} for optimisation and SGLang \citep{zheng2024sglang} for rollout generation. Both the warm-up and downstream fine-tuning use GRPO \citep{shao2024deepseekmath} with $8$ rollouts per task, full-parameter updates and Adam at a learning rate of $1\times10^{-6}$ with $\beta = (0.9, 0.98)$ and weight decay $0.01$, a clipped surrogate with $\epsilon = 0.2$ and $\epsilon_{\text{high}} = 0.28$, truncated importance sampling clipped to $[0.5, 1.5]$, a low-variance KL loss against the initial policy with coefficient $0.01$, and an entropy coefficient of $0.001$. Rollouts are sampled at temperature $1.0$ with at most $8192$ response tokens per action turn. The warm-up draws its $1600$ tasks at random in a single pass and takes $4$ optimiser updates of batch size $32$ per step. Downstream fine-tuning takes $8$ labelled examples and one optimiser update of batch size $64$ per step, for $n_{\max}/8$ steps, where $n_{\max}$ is the largest training budget of the benchmark. The two models on a curve see the same examples in the same order and share the seed, limits, decoding and scorer, so they differ only in their initial weights; evaluation uses the official scorer on the full held-out set.

\paragraph{Evaluation metrics.}
Following standard practice in self-supervised pre-training, we evaluate the warm-up stage by measuring the labelled downstream data required to reach a given performance level~\citep{henaff2020cpc,chen2020simclr,chen2020simclrv2}. We define downstream label efficiency as the number $n$ of annotated examples required to reach a given accuracy. For each label-efficiency curve, we report two statistics. The \emph{curve-level gain} is the mean accuracy difference between \ours{} and the baseline across all training-data budgets and serves as our primary comparison. The \emph{matching budget} is the smallest baseline budget needed to match or exceed \ours{}'s accuracy, expressed as a multiple of the corresponding \ours{} budget. It summarises the low-budget regime; ``never ($>k\times$)'' indicates that the baseline does not reach the target within the evaluated range.

\section{Results}
\label{sec:main}

\subsection{Agentic data analysis}
\label{sec:eff}

Figure~\ref{fig:main} gives an overview of the downstream fine-tuning curves, and Table~\ref{tab:main} reports the detailed curve-level gain and matching budget. Each curve averages three independent runs, with shaded bands showing their minimum and maximum across runs. Across two models and six benchmarks, \ours{} enables agentic LLMs to reach or even surpass baseline performance with substantially fewer annotated training examples. The advantage is most pronounced when labels are scarce.
Notably, the warm-up tasks differ in form from the downstream ones. The warm-up trains the model to reconstruct synthetic trajectories, whereas the downstream tasks span data science coding, report analysis, table question answering, and other related data analysis tasks. Nevertheless, the performance gap generally persists as fine-tuning proceeds. On TableBench at $9$B, although the \ours{} model starts below the baseline, it overtakes the baseline after only $16$ examples.
In terms of label efficiency, on every curve, the baseline requires more annotated examples to match the accuracy achieved by the \ours{} model at its smallest training budget. Specifically, the baseline requires between $8\times$ and $32\times$ as many examples when it reaches the target, with a median of $16\times$; on the two DSEval curves, it does not reach the target within the evaluated range.

Recent RL studies show that repeatedly training on a single example can keep improving agentic LLMs \citep{wang2025rlone,li2026onesample}, as a well-designed example leads the model through different intermediate states across iterations~\citep{fu2026rethinking,liuposition}. We test this extreme low-label regime and hold one annotated example fixed for both models and train on it for up to $32$ steps. An example qualifies only if both models show a non-zero spread of rewards within its group, so that both are learning from it, and among the qualifying examples we take those at the $25$th, $50$th and $75$th percentile of the per-group accuracy of the baseline. Figure~\ref{fig:oneshot} reports the accuracy difference after each step, averaged over the six benchmarks. After $32$ steps on the same example, the \ours{} model still leads by $5.5$\% to $12.6$\%, and the average difference stays positive at every step and every difficulty.

\begin{table}[t]
\centering
\caption{Results on six agentic data-analysis benchmarks. \emph{EnvLunch}/\emph{Baseline} denote the mean accuracy averaged across all training budgets on each curve; $\Delta$ is their difference. Each value is the mean over three independent fine-tuning runs, with the standard deviation across runs shown as a subscript ($\Delta$ is computed by pairing runs). \emph{Match} denotes the matching budget defined in Section~\ref{sec:setup}.}
\label{tab:main}
\resizebox{\linewidth}{!}{\input{tables/main}}
\end{table}

\subsection{Beyond agentic data analysis}
\label{sec:cross}

We extend the \ours{} model to seven downstream benchmarks beyond agentic data analysis including text-to-SQL (BIRD, \citealp{li2023bird}; Spider, \citealp{yu2018spider}; Spider2-lite, \citealp{lei2024spider2}), question answering (HotpotQA, \citealp{yang2018hotpotqa}), embodied simulation (ALFWorld, \citealp{shridhar2021alfworld}), application use (AppWorld, \citealp{trivedi2024appworld}), and spreadsheet editing (SpreadsheetBench, \citealp{ma2024spreadsheetbench}).
On BIRD, Spider and the local SQLite subset of Spider2-lite, each turn either runs a read-only SQL query, whose rows or error message is returned, or submits a final query, scored by execution accuracy. HotpotQA uses the distractor setting and proportional match. ALFWorld takes text commands, returns the game feedback with the list of admissible commands and is scored by success rate. In AppWorld the agent writes Python that calls the application APIs and ends the episode by calling the task-completion API, scored by task goal completion. In SpreadsheetBench the agent writes \texttt{openpyxl} code that produces an output workbook, may inspect a recalculated preview of it before submitting, and is scored by execution accuracy. The held-out sizes and turn limits are $1534$ and $32$ for BIRD, $1034$ and $32$ for Spider, $61$ and $8$ for Spider2-lite, $400$ questions sampled from the validation split and $32$ for HotpotQA, the $134$ unseen games and $30$ for ALFWorld, $57$ and $30$ for AppWorld, and $394$ and $6$ for SpreadsheetBench. Training budgets range over $16$--$2048$ examples for BIRD and Spider, $8$--$64$ for Spider2-lite and AppWorld, $8$--$256$ for SpreadsheetBench, and $16$--$1024$ for the rest.
Table~\ref{tab:cross} reports the curve-level gain and the matching budget on these seven benchmarks. For these tasks, the warm-up differs entirely from the downstream tasks in task formulation and domain, yet \ours{} still exhibits a degree of generalisation, improving downstream label efficiency. Although the free lunch claimed by \ours{} is in the form of additional data rather than reduced compute, across the data analysis and additional benchmarks, the extra fine-tuning required for the baseline to match \ours{} at its smallest budget takes longer than the warm-up stage: summed over the thirteen benchmarks, it takes about $7$ hours at $9$B and $4$ hours at $4$B on the same GPU node (a lower bound where the baseline never reaches the target), against a one-off warm-up of about $4$ and $3$ hours.

\begin{table}[t]
\centering
\begin{minipage}[t]{0.60\linewidth}
\vspace{0pt}
\centering
\begin{minipage}[t][\sbcap][t]{\linewidth}
\caption{Seven benchmarks beyond data analysis. $\Delta$ is the
curve-level gain, \emph{Match} the matching budget of Section~\ref{sec:setup}.}
\label{tab:cross}
\end{minipage}
\footnotesize
\setlength{\tabcolsep}{4pt}
\input{tables/crossdomain_compact}
\end{minipage}\hfill
\begin{minipage}[t]{0.37\linewidth}
\vspace{0pt}
\makeatletter\def\@captype{figure}\makeatother
\centering
\includegraphics[width=\linewidth]{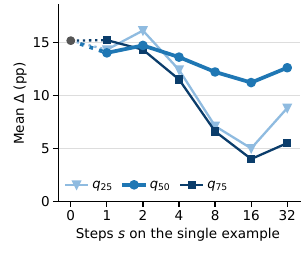}
\caption{One labelled example, $9$B. Mean $\Delta$ after $s$ steps on
a \emph{single} fixed example; $q_k$: $k$th-percentile example.}
\label{fig:oneshot}
\end{minipage}
\end{table}

\section{Analysis of the warm-up}
\label{sec:mech}

We conduct a detailed analysis of the trajectories and design additional controlled experiments to understand what the model learns from the \ours{} warm-up and how it shapes the model's behaviour, enabling it to make better use of downstream data.
We first examine four intuitive alternative explanations, then analyse and quantify the characteristics of the trajectories under a fair setting, and finally measure how the model reduces its uncertainty about the environment in a controlled experimental setting.

\paragraph{Alternative Explanations.}
\label{sec:subject}

\begin{figure}[t]
\centering
\begin{minipage}[t]{0.51\linewidth}
\centering
\includegraphics[height=\halffigheight]{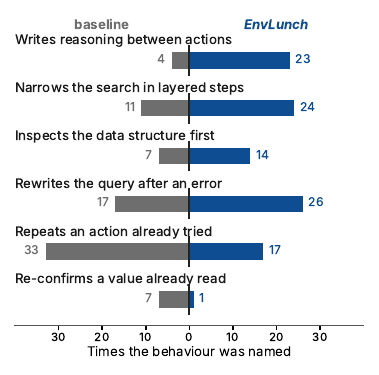}
\caption{Behaviours identified in trajectories, where both models obtain the same score.}
\label{fig:traj}
\end{minipage}\hfill
\begin{minipage}[t]{0.45\linewidth}
\centering
\includegraphics[height=\halffigheight]{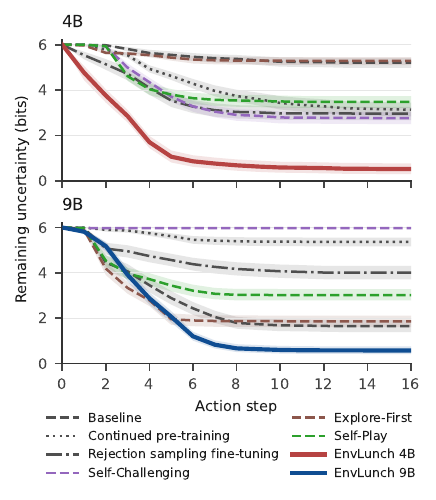}
\caption{Remaining uncertainty about the environment after each action, with 95\% intervals over tasks.}
\label{fig:uncert}
\end{minipage}
\end{figure}

\ours{} does not ask the model to learn an explicit world model of the environment. Instead, the task simply requires the model to interact with and reason about the environment in order to reconstruct a randomly sampled trajectory. What, then, does the model actually learn during this RL process? We consider four intuitive possibilities.
First, the model is unlikely to benefit simply by memorising the warm-up environments, since none of them appears in any evaluation set (Section~\ref{sec:setup}), and benchmarks whose domains do not overlap with them, such as ALFWorld and AppWorld, still benefit from \ours{}. Two other possibilities are that the model learns the schema of the warm-up environments, or that it imitates some key actions appearing in successful trajectories. We test these hypotheses by replacing the RL-based warm-up with continued pre-training on environment text and rejection sampling fine-tuning on successful trajectories, respectively. For the former, a template sampler writes, for each table of the dataframe corpus, documents giving the file name, the columns and their types, a templated \texttt{pandas} expression and its execution result ($34{,}267$ documents, $6.60$M tokens), and we train on them for one epoch of full-parameter next-token prediction. For the latter, we sample up to $8$ attempts from the backbone on each of the $1600$ \ours{} tasks under the same limits and scoring, keep the attempts with full reward, and fine-tune on them for one epoch with the loss on model tokens only. Neither of these two warm-up approaches yields a gain: averaged over the six data-analysis benchmarks, their curve-level gains over the baseline are $-3.5$ and $-5.7$ at $4$B and $-2.4$ and $-4.7$ at $9$B, respectively.
A fourth possibility is that any reinforcement learning on these environments brings the gain, whatever the tasks are. We test it by keeping the \ours{} stage unchanged and replacing only its task pool with tasks that the LLM synthesises itself over the same databases and tables, following three lines of prior work. In Self-Challenging \citep{zhou2025selfchallenging}, the model inspects an environment (field descriptions, table schemas or creation statements, and sample data) and writes a question together with a program that answers it as verification, and tasks whose program fails are dropped. In Explore-First \citep{sun2025osgenesis,su2025learnbyinteract}, the model first interacts with the environment freely and then rewrites its trajectory into a question and an answer. In Self-Play \citep{zhao2025absolutezero,huang2025rzero}, the model proposes a question and solves it eight times, taking the majority answer as the label, and tasks are dropped when all eight attempts agree or fewer than three agree. In all three, a backbone of the same scale generates the same number of tasks over the same environments, and the harness, tools, interaction limits and all other settings are shared with \ours{}. At both scales, all three fall short of \ours{} on the downstream curves: their curve-level gains are $+5.2$, $+5.3$ and $+6.4$ at $4$B and $+1.6$, $-0.3$ and $-4.0$ at $9$B, against $+9.6$ and $+9.3$ for \ours{}. In each of them the model labels its own tasks, and nothing checks that the label answers the question it wrote, which likely makes the training signal less reliable.

\paragraph{Trajectory-Level Behavioural Analysis.}
\label{sec:traj}

\begin{figure}[t]
\centering
\includegraphics[width=\linewidth]{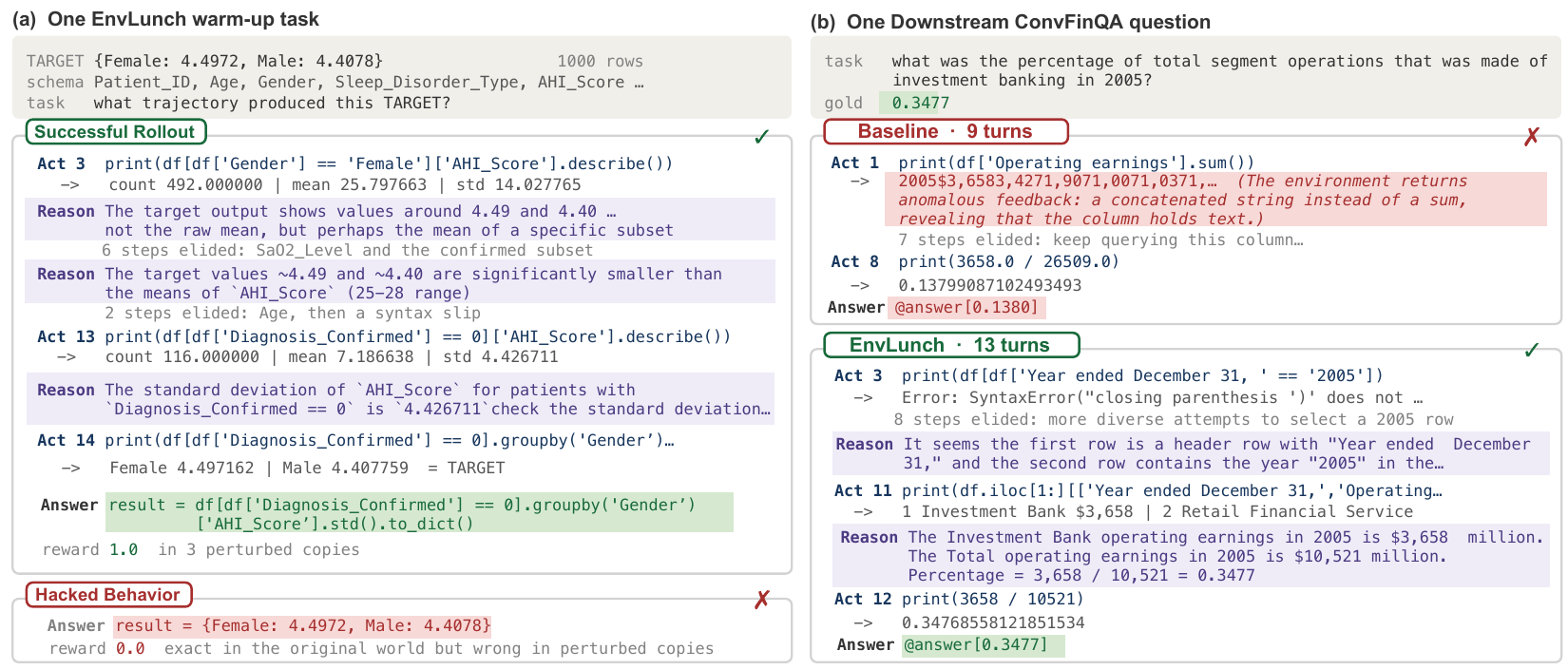}
\caption{Two excerpted trajectories from the $9$B models. \textbf{(a)} An \ours{} warm-up task. \textbf{(b)} A ConvFinQA question after both models have been fine-tuned on the same labelled examples.}
\label{fig:case}
\end{figure}

We randomly sample trajectories from the baseline model and the model after \ours{} warm-up across all benchmarks, and retain only items on which both models obtain the same score before fine-tuning, ensuring a fair comparison ($78$ items). We present each pair without scores, with the two models labelled \emph{A} and \emph{B} by a hash of the item name so that their order is shuffled, and ask Claude Sonnet 5 to describe in free text the behaviours in which the two trajectories differ, with no taxonomy supplied. We then manually verify the descriptions against the trajectories, and categorise and quantify the identified differences. As shown in Figure~\ref{fig:traj}, after \ours{} warm-up, the model more often reasons explicitly at key intermediate steps, more frequently inspects the data structure before executing an operation, progressively narrows down the search space, and rewrites the query after encountering errors. In contrast, the baseline more frequently repeats previously attempted operations or reconfirms values that it has already retrieved.
Figure~\ref{fig:case} presents a concrete case study. In the \ours{} stage in panel (a), the model observes that the values in the final observation are substantially smaller than the mean returned by its query. It therefore reasons further and ultimately obtains the correct answer. In the downstream ConvFinQA task in panel (b), although the task type is different, the model exhibits a similar pattern of reasoning and consequently takes more appropriate actions.
Next, we investigate what kind of environment interaction behaviour this reasoning shapes.

\paragraph{Uncertainty Reduction in a Controlled Environment.}
\label{sec:probeworld}

As shown by the trajectory analysis above, after warm-up, the model appears to learn how to interact with the environment to reduce uncertainty and narrow the search space for subsequent steps.
However, real-world benchmark environments are large and complex, making it difficult to cleanly and accurately measure how well a model understands the environment.
Following prior work that measures information seeking by how strongly an agent narrows hypotheses about a hidden state \citep{grand2024loselips,hu2024uncertainty,golde2025mastermindeval}, we construct a controlled environment with exactly measurable uncertainty. The model queries a constructed sales table whose three columns each admit one of four encodings, yielding $64$ possible environments. All environments share the same prompt and final answer and differ only in their hidden encodings, which can be distinguished only through queries.
Concretely, the model is asked for the total amount of the completed orders of one product in one target town, given the column documentation and the first rows of a table covering $20$ towns. The records of the target town write the town, status and product columns in one of four ways each (e.g., the town as \texttt{Marlborough}, \texttt{MLB}, \texttt{Mrlbr} or \texttt{Mboro}). The documentation warns that some records use legacy codes without naming them, a common data-quality problem \citep{rahm2000datacleaning,kim2003dirtydata}, and the model never works in the version that matches the documentation, so it must inspect the data to compute the answer.
Following possible-world semantics for incomplete databases \citep{imielinski1984incomplete}, we execute each model action across all $64$ hidden tables and discard those whose output differs from the model's observation. The remaining tables are consistent with the model's observations, so we define uncertainty as $\log_2$ of their number, decreasing from $6$ to $0$ bits as the encodings are identified. Six domains contribute $25$ tasks each. The models are evaluated right after their warm-up, without downstream fine-tuning, in the DataBench harness with at most $24$ turns; each task is run once with greedy decoding and twice with sampling (temperature $0.7$, top-$p$ $0.8$), and the $95\%$ intervals come from bootstrap resampling over tasks.
Figure~\ref{fig:uncert} shows that \ours{} reduces uncertainty more effectively than the baseline and the continued pre-training, rejection sampling fine-tuning and synthesised-task variants at both scales. This difference is not explained by taking more actions: at $4$B, \ours{} and rejection sampling fine-tuning take a similar number of actions but achieve substantially different uncertainty reduction. The warm-up therefore shapes the model to select actions that reduce uncertainty about the environment, providing a behavioural mechanism for the downstream gains in Section~\ref{sec:main}.

\section{Conclusion}
\label{sec:discussion}

We propose \ours{}, a warm-up stage that trains agentic LLMs to reconstruct randomly synthesised trajectories from the environment alone before downstream agentic reinforcement learning. \ours{} improves accuracy by over $9$\% on average across six agentic data analysis benchmarks at both model scales. When labels are scarce, the baseline requires $8$--$32\times$ as many annotated examples to reach comparable performance, and the gains transfer to almost all of the seven additional benchmarks. Our study focuses on relational databases and dataframes with verifiable outcomes at the $4$B and $9$B scales, and extending the warm-up to other executable environments, open-ended tasks and larger models is left for future work. More broadly, these results show that an environment's existing information and executable dynamics can provide useful supervision for learning agent behaviour before any downstream question is written.

\section*{Acknowledgements}
This work was supported in part by the UK Engineering and Physical Sciences Research Council (EPSRC) through a Turing AI Fellowship (grant no. EP/V020579/1, EP/V020579/2).

\bibliographystyle{iclr2027_conference}
\bibliography{refs}

\end{document}

%% file: tables/main.tex
\begin{tabular}{llrrrrrrrr}
\toprule
& & \multicolumn{4}{c}{9B} & \multicolumn{4}{c}{4B} \\
\cmidrule(lr){3-6}\cmidrule(lr){7-10}
Benchmark & Data substrate & \ours{} & Baseline & $\Delta$ & Match & \ours{} & Baseline & $\Delta$ & Match \\
\midrule
ConvFinQA & Table + text & $77.9_{\pm 0.2}$ & $68.2_{\pm 2.0}$ & $\mathbf{+9.8}_{\pm 2.1}$ & 32$\times$ & $75.9_{\pm 0.6}$ & $67.8_{\pm 2.2}$ & $\mathbf{+8.1}_{\pm 1.8}$ & 8$\times$ \\
TAT-QA & Table + text & $88.7_{\pm 1.3}$ & $75.9_{\pm 1.5}$ & $\mathbf{+12.8}_{\pm 0.7}$ & 8$\times$ & $89.6_{\pm 0.3}$ & $78.9_{\pm 1.3}$ & $\mathbf{+10.7}_{\pm 1.4}$ & 8$\times$ \\
MultiHiertt & Table + text & $27.1_{\pm 3.1}$ & $15.7_{\pm 0.7}$ & $\mathbf{+11.4}_{\pm 3.1}$ & 32$\times$ & $38.9_{\pm 1.6}$ & $24.7_{\pm 3.7}$ & $\mathbf{+14.3}_{\pm 4.0}$ & 16$\times$ \\
TableBench & Table & $66.4_{\pm 1.4}$ & $59.1_{\pm 2.5}$ & $\mathbf{+7.3}_{\pm 3.6}$ & 8$\times$ & $62.3_{\pm 0.4}$ & $53.1_{\pm 1.4}$ & $\mathbf{+9.2}_{\pm 1.6}$ & 16$\times$ \\
DSEval & DataFrame & $70.2_{\pm 3.0}$ & $59.3_{\pm 1.7}$ & $\mathbf{+10.9}_{\pm 3.1}$ & never & $60.8_{\pm 2.5}$ & $53.6_{\pm 1.2}$ & $\mathbf{+7.1}_{\pm 3.6}$ & never \\
DataBench & DataFrame & $80.2_{\pm 1.2}$ & $76.4_{\pm 1.1}$ & $\mathbf{+3.8}_{\pm 2.0}$ & 16$\times$ & $76.4_{\pm 2.5}$ & $68.0_{\pm 1.6}$ & $\mathbf{+8.4}_{\pm 3.7}$ & 32$\times$ \\
\midrule
\multicolumn{2}{l}{\emph{Mean over 6 benchmarks}} &  &  & \textbf{+9.32} &  &  &  & \textbf{+9.63} &  \\
\bottomrule
\end{tabular}

%% file: tables/crossdomain_compact.tex
\begin{tabular}{lrcrc}
\toprule
& \multicolumn{2}{c}{9B} & \multicolumn{2}{c}{4B} \\
\cmidrule(lr){2-3}\cmidrule(lr){4-5}
Benchmark & $\Delta$ & Match & $\Delta$ & Match \\
\midrule
BIRD & \textbf{+4.8} & 64$\times$ & \textbf{+6.8} & 64$\times$ \\
Spider & \textbf{+2.2} & 32$\times$ & \textbf{+1.7} & 4$\times$ \\
Spider2-lite & \textbf{+4.6} & never & \textbf{+1.4} & 1$\times$ \\
\addlinespace[2pt]
HotpotQA & \textbf{+15.3} & never & \textbf{+14.0} & 32$\times$ \\
ALFWorld & \textbf{+3.7} & 8$\times$ & \textbf{+2.1} & 4$\times$ \\
AppWorld & \textbf{+5.7} & never & \textbf{+0.4} & 1$\times$ \\
SpreadsheetBench & \textbf{+2.0} & never & $-$0.2 & 1$\times$ \\
\midrule
\emph{Mean} & \textbf{+5.48} &  & \textbf{+3.73} &  \\
\bottomrule
\end{tabular}